\documentclass{article}

\usepackage{microtype}
\usepackage{graphicx}
\usepackage{caption}
\usepackage{subcaption}
\usepackage{booktabs} 

\usepackage{hyperref}

\usepackage[accepted]{icml2023}

\newif\ifshowtmp
\showtmptrue
\ifshowtmp

\else

\fi

\newcommand{\Creftodo}[1]{\todo{cref}}
\newcommand{\creftodo}[1]{\todo{cref}}
\newcommand{\citetodo}[1]{\todo{cite}}

\newcommand{\Creftd}[1]{\todo{cref}}
\newcommand{\creftd}[1]{\todo{cref}}
\newcommand{\citetd}[1]{\todo{cite}}

\usepackage{amsmath}
\usepackage{amssymb}
\usepackage{mathtools}
\usepackage{amsthm}
\usepackage{xcolor}

\usepackage[capitalize,noabbrev]{cleveref}

\theoremstyle{plain}

\theoremstyle{definition}

\theoremstyle{remark}

\usepackage[textsize=tiny]{todonotes}

\newcommand{\Mname}{Action-Free Guide}
\newcommand{\Mabbr}{AF-Guide}

\newcommand\blfootnote[1]{%
  \begingroup
  \renewcommand\thefootnote{}\footnote{#1}%
  \addtocounter{footnote}{-1}%
  \endgroup
}

\icmltitlerunning{Guiding Online Reinforcement Learning with Action-Free Offline Pretraining}

\begin{document}

\twocolumn[
\icmltitle{Guiding Online Reinforcement Learning with Action-Free Offline Pretraining}

\icmlsetsymbol{equal}{*}

\begin{icmlauthorlist}
\textbf{Deyao Zhu}\textsuperscript{1}~~
\textbf{Yuhui Wang}\textsuperscript{1}~~
\textbf{J\"uergen Schmidhuber}\textsuperscript{1,2,3}~~
\textbf{Mohamed Elhoseiny}\textsuperscript{1}~~
\\
\textsuperscript{1}King Abdullah University of Science and Technology\\
\textsuperscript{2}The Swiss AI Lab IDSIA/USI/SUPSI~~
\textsuperscript{3}NNAISENSE\\
\texttt{\{deyao.zhu, yuhui.wang, juergen.schmidhuber, mohamed.elhoseiny\}@kaust.edu.sa}
\end{icmlauthorlist}

\icmlcorrespondingauthor{Deyao Zhu}{Deyao.Zhu@kaust.edu.sa}

\icmlkeywords{Machine Learning, ICML}

\vskip 0.3in
]




\begin{abstract}

Offline RL methods have been shown to reduce the need for environment interaction by training agents using offline collected episodes. However, these methods typically require action information to be logged during data collection, which can be difficult or even impossible in some practical cases.
In this paper, we investigate the potential of using action-free offline datasets to improve online reinforcement learning, name this problem Reinforcement Learning with Action-Free Offline Pretraining (AFP-RL).
We introduce \Mname~(\Mabbr),
a method that guides online training by extracting knowledge from action-free offline datasets.
\Mabbr~consists of an Action-Free Decision Transformer (AFDT) implementing a variant of Upside-Down Reinforcement Learning. It learns to plan the next states from the offline dataset, and a Guided Soft Actor-Critic (Guided SAC) that learns online with guidance from AFDT.
Experimental results show that \Mabbr~can improve sample efficiency and performance in online training thanks to the knowledge from the action-free offline dataset. 
Code is available at \url{https://github.com/Vision-CAIR/AF-Guide}.

\blfootnote{Preprint}
\end{abstract}

\section{Introduction}
\label{intro}

Training a reinforcement learning agent directly in the environment from scratch can be a challenging task, as it usually requires a large number of time-consuming and costly interaction steps to explore the environment. 
As a result, improving sample efficiency in RL has become one of the most important directions in the reinforcement learning community.
Offline reinforcement learning methods use offline collected episodes only to train RL agents.
After the offline RL training, RL agents can be further online finetuned in the environment with much fewer interaction steps.
To apply offline RL methods, actions need to be logged when collecting the offline episodes. 
However, recording actions (e.g., motor torques) can be difficult or even impossible in certain practical cases, like learning from large-scale internet videos.
Despite the lack of actions, these data (e.g., a video of a robot making a pizza) still hold valuable information about agents' movements and environments' transitions, which can be utilized to instruct the RL agent on the distinctions between advantageous and detrimental movements.

\begin{figure*}
    \centering
    \includegraphics[width=0.85\textwidth]{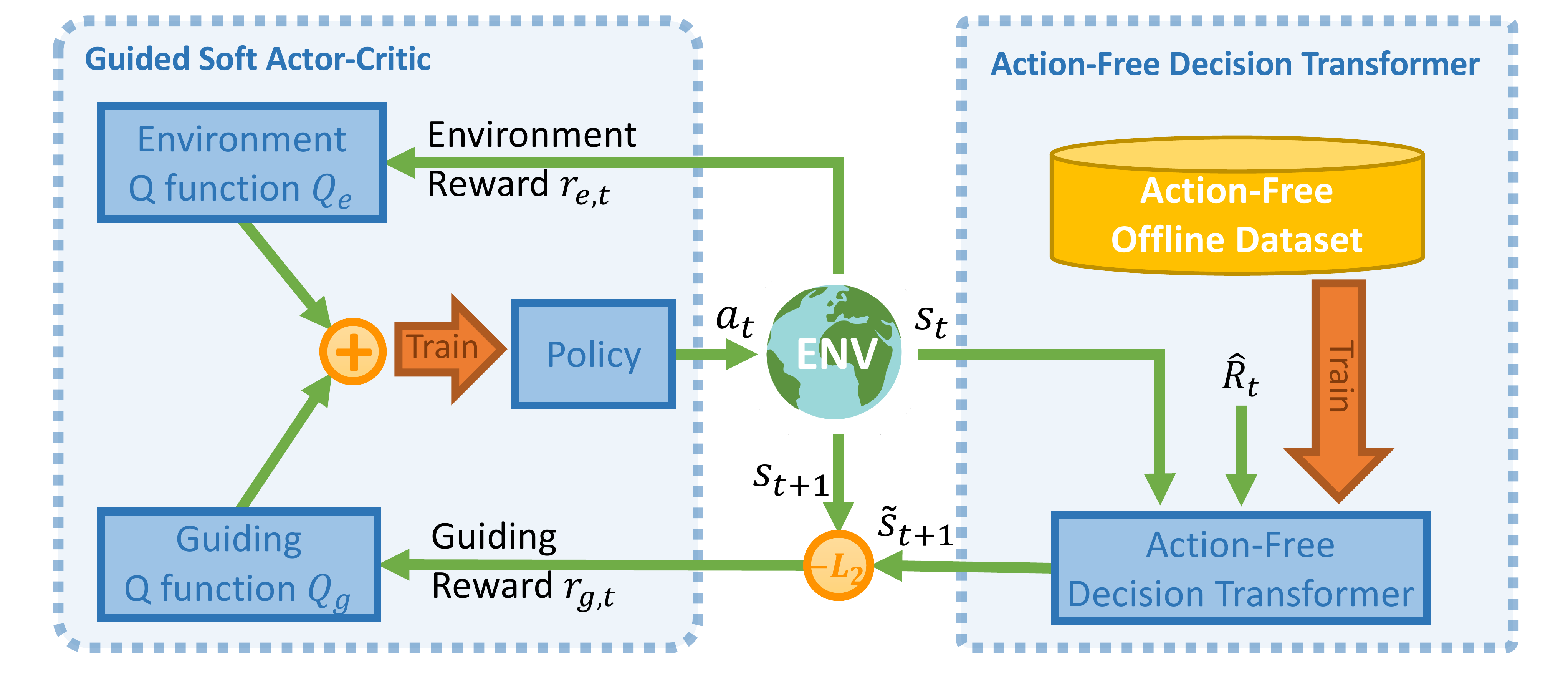}
    \caption{An overview of \Mabbr. Action-Free Decision Transformer (AFDT) is trained on the action-free offline dataset to plan the next state $\tilde s_{t+1}$ given previous states and the desired return-to-go $\hat R_t$. The guiding reward $r_g$ is formed based on the negative L2 distance between the planned state $\tilde s_{t+1}$ and the real state $s_{t+1}$. In addition to SAC's original Q function denoted as $Q_e$ that fits the environment reward $r_e$, Guided SAC has an additional Q function $Q_g$ to fit the guiding reward $r_g$ with zero discount factor to discard the future return.
    The policy is trained by the weighted sum over the two Q functions.}
    \label{fig:overview}
\end{figure*}

In this paper, we explore the potential of utilizing action-free offline reinforcement learning datasets to guide online Reinforcement Learning. 
We name this setting Reinforcement Learning with Action-Free Offline Pretraining (AFP-RL).
We propose \Mname~(\Mabbr), a method that improves online training by learning to plan good target states from the action-free offline dataset.
\Mabbr~comprises two main components that we introduce: an Action-Free Decision Transformer (AFDT), and a Guided Soft Actor-Critic (Guided SAC). 
AFDT, a variant of the Upside Down Reinforcement Learning \cite{schmidhuber2019reinforcement} model Decision Transformer \cite{chen2021decision}, is trained on an offline dataset without actions to plan the next states based on the past states and the desired future returns. 
Guided SAC, a variation of SAC \cite{haarnoja2018soft, haarnoja2018soft_2}, follows the planning of AFDT by maintaining an additional Q function that fits an intrinsic reward built from the discrepancy between the planned state and the achieved state with zero discount factor. 
An overview of our method is summarized in Fig.\ref{fig:overview}.
Our experimental results demonstrate that \Mabbr~can significantly improve sample efficiency during online training by utilizing action-free offline datasets.

Our contribution can be summarized as follows:

\begin{itemize}
    \item We propose Reinforcement Learning with Action-Free Offline Pretraining (AFP-RL), a novel setting to study how to guide online Reinforcement Learning with action-free offline datasets.
    
    \item We present \Mname~(\Mabbr), a method that 
    pretrains a model which can extract knowledge from the action-free offline dataset and conduct state-space planning to guide online policy learning.

    \item Experimental results show that \Mabbr~can benefit from the action-free offline dataset to improve sample efficiency and performance during online training. 
\end{itemize}

\section{Related Work}
\paragraph{Offline Reinforcement Learning}
Offline reinforcement learning methods learn policies using pre-collected episodes from unknown behavior policies. 
Many offline RL methods, such as
CQL \cite{kumar2020conservative}, IQL \cite{kostrikov2021offline_2}, AWAC \cite{nair2020awac}, BCQ, and COMBO \cite{yu2021combo}, have been developed from off-policy algorithms, with additional constraints to avoid out-of-distribution actions that are not covered by the dataset.
Recently, Decision Transformer \cite{chen2021decision} and Trajectory Transformer \cite{janner2021offline} convert the offline RL problem as a context-conditioned sequential modeling problem and generate good actions by either conditioning on desired future return following the Upside Down Reinforcement Learning framework \cite{schmidhuber2019reinforcement} or searching for a good rollout with a high future return.
In our AFP-RL setting, datasets do not contain explicitly labeled actions (although the dataset may contain videos of acting agents).
In this case, learning an offline policy directly is infeasible. 
Our method AFDT-Guide instead leverages action-free data to plan good target states and guide online training for improved performance.

\paragraph{Imitation Learning from Observation}
The target of imitation learning from observation is to learn a policy through state-only action-free demonstrations from experts.
imitation Learning from observation methods can be broadly classified into different categories.
Methods like GSP \cite{pathak2018zero} and BCO \cite{torabi2018behavioral}
train an inverse dynamic model to infer the expert actions given state transitions.
Intrinsic-reward-based methods like DeepMimic \cite{peng2018deepmimic},  Context-Aware Translation \cite{liu2018imitation}, and \citet{lee2021generalizable}, 
create surrogate reward functions to guide the online training.
Other methods like GAIfO \cite{torabi2018generative}, IDDM \cite{yang2019imitation}, MobILE \cite{kidambi2021mobile} employ adversarial learning.
The difference between imitation learning from observation and our setting AFP-RL is similar to the difference between imitation learning and offline RL.
In imitation learning from observation, the dataset is collected by an expert policy, and agents are trained to directly imitate the collected episodes.  
In contrast, episodes in AFP-RL are collected by behavior policies that may be suboptimal. As a result, directly imitating these episodes would lead to suboptimal performance.

\paragraph{Motion Forecasting}
Motion forecasting is the task of predicting the future motion of agents given past and context information.
It helps autonomous systems like autonomous driving and robotics to foresee and avoid potential risks like collisions in advance.
Recent methods for motion forecasting have explored various architectural designs.
For example, Social-LSTM \cite{alahi2016social} and Trajectron++ \cite{salzmann2020trajectron++} are based on RNN. Social-GAN \cite{gupta2018social} and HalentNet \cite{zhu2021halentnet} benefit from generative adversarial training. 
Social-STGCNN \cite{mohamed2020social} and Social-Implicit \cite{mohamed2022social} predict the future via spatial-temporal convolution.
AgentFormer \cite{yuan2021agentformer}, mmTransformer \cite{liu2021multimodal}, and ST-Transformer \cite{aksan2021spatio} are models based on Transformer architecture \cite{vaswani2017attention} designed for pedestrian or vehicle trajectory prediction.
Our state-planner AFDT is a Transformer model.
Instead of simply predicting the future states conditioned on the past, AFDT plans the future states by additionally conditioning on the desired future return in the UDRL framework \cite{schmidhuber2019reinforcement}.

\section{Background}


\paragraph{Soft Actor-Crtic (SAC) } 
SAC is an actor-critic RL approach based on the maximum entropy framework \cite{haarnoja2018soft, haarnoja2018soft_2}, which involves optimizing a Q network $Q_e$ \footnote{We use the subscript $e$ to denote notations related to the {e}nvironment reward, and will use $g$ to differentiate the notations related to the {g}uiding reward (see \Cref{sec:gsac}).} and the policy network $\pi$.
The Q function $Q_{e}$ is learned with the following objective 
\begin{equation}\label{eq_Q_SAC}
\min_{Q_{e}}\mathbb{E}_{\mathcal{D}_{\mathrm{online} }}\Vert Q_{e} (s_{t} ,a_{t} )-
Q_{e,t}^{\rm target}
\Vert_2^2
\end{equation}
where $\mathcal{D}_{\mathrm{online}} \triangleq \{ (s_t,a_t, r_{e,t}, s_{t+1} ) \} $ with state $s_t$, action $a_t$, environment reward $r_{e,t}$, and next state $s_{t+1}$, is the online replay buffer. 
$Q_{e,t}^{\rm target}$ is the target Q value computed as follows
\begin{footnotesize}
\begin{equation}
\label{eq_Q_target_SAC}
 Q_{e,t}^{\rm target} = r_{e,t}+\gamma \mathbb E_{ \pi} \left[Q_{e}(s_{t+1}, a_{t+1}) - \alpha \log \pi(a_{t+1}|s_{t+1}) \right]    
\end{equation}
\end{footnotesize}

Here, $\gamma$ is the discount factor and $\alpha$ is the temperature parameter to weight the entropy.
The policy network is learned by the following objective
\begin{equation}
\label{eq_sac_policy}
\min_{\pi} \mathbb{E}_{s_t\sim\mathcal{D}_{\mathrm{online} }, a_t \sim \pi }
\left[
    \alpha \log( \pi( a_t|s_t ) ) -Q_e(s_t,a_t)
\right]
\end{equation}


\paragraph{Upside Down Reinforcement Learning and Decision Transformer}
Traditional Reinforcement Learning methods are trained to predict future rewards (e.g., a Q function) first and convert the prediction into rewarding actions. 
In contrast, Upside Down Reinforcement Learning (UDRL)  \cite{schmidhuber2019reinforcement} framework takes desired future rewards as inputs to generate actions.
As an instance of UDRL in the offline RL setting, Decision Transformer (DT) \cite{chen2021decision} is trained in the offline dataset to regress the current action $a_t$ conditioned on the past $K$ states $s_{t-k:t}$, actions $a_{t-k:t-1}$, and the future returns (named Return-To-Go, RTG) $\hat R_{t-k:t}$, with $\hat R_t = \sum_{t'=t}^T r_{t'}$. 
The architecture of DT is based on the language model GPT \cite{radford2018improving}.
When evaluated in an environment, the model is provided with an initial state $s_0$ and a desired initial RTG $\hat R_0$ to generate the first action.
After executing the action $a_t$ in the environment and observing the reward $r_t$ and the next state $s_{t+1}$, the RTG is updated by $\hat R_{t+1} = \hat R_{t} - r_{t}$. 
The executed action $a_t$, the current return-to-go $\hat R_{t+1}$, and the current state $s_{t+1}$ are then fed back into DT to infer the next action.
Given a high initial RTG $\hat R_0$, DT is able to generate good actions that lead to high future returns.
Due to the dependence of standard DT on action labels, it can not be directly applied for Action Free Pretraining.


\section{\Mname}

\paragraph{Action-Free Offline Pretraining}

In the setting of Reinforcement Learning with Action-Free Offline Pretraining (AFP-RL), 
an action-free offline dataset, $\mathcal{D} = \{ \tau_1, \tau_2, ..., \tau_N \}$, is provided to boost the online training in the environment.
The trajectories in the dataset have been pre-collected in the environment by behavior policies that are unknown to the agent. 
Each trajectory, $\tau$, contains states and rewards in the format $\tau = (s_0, r_0, s_1, r_1, ..., s_T, r_T)$, with $T$ time steps. 
Unlike traditional offline RL, where the policy is learned directly from the offline dataset, it is infeasible to learn a policy from an action-free offline dataset as it lacks the necessary action information. 
However, such a dataset still contains valuable information about the agent's movements and the environment's dynamics. 
Our proposed setting, AFP-RL, aims to leverage this information to improve online training.

\paragraph{Methodology Overview}
Our method, \Mname~ (\Mabbr), utilizes knowledge from action-free offline datasets by training an Action-Free Decision Transformer (AFDT) on these datasets to plan the next states that lead to high future returns. 
Then, the online agents, trained by Guided Soft Actor-Critic (Guided SAC), follow the planning with an additional Q function optimized for an intrinsic reward based on the planned states. 
\Mabbr~ is similar to the Learning-to-Think framework \cite{schmidhuber2015learning, schmidhuber2018one}: Guided SAC sends queries (in this case desired future
returns) into AFDT and learns to use the answers (in this case good next states) to improve its own performance. 
The overall methodology is illustrated in Fig.\ref{fig:overview}.

\subsection{Action-Free Decision Transformer}

\begin{figure}[t]
\vskip 0.2in
\begin{center}
\centerline{\includegraphics[width=\columnwidth]{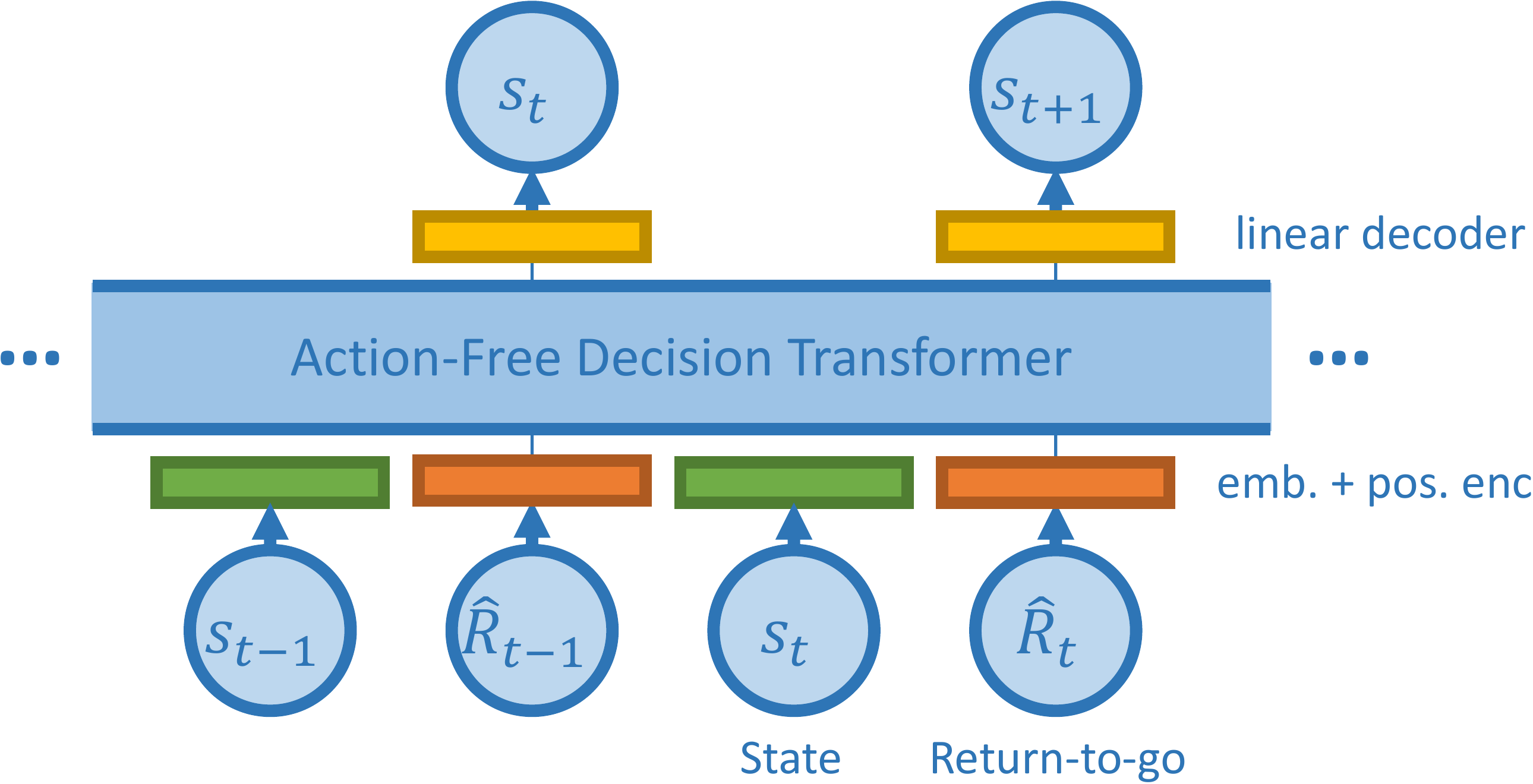}}
\caption{Action-Free Decision Transformer. The next state is planned given previous states and a desired future return, named return-to-go.}
\label{fig:afdt}
\vskip -0.3in
\end{center}
\end{figure}

\Mname (\Mabbr) can be considered as a variant of the UDRL model Decision Transformer (DT) \cite{chen2021decision} that we designed to operate on action-free offline datasets.
Unlike DT, which predicts actions based on past RTGs, states, and actions, AFDT plans the next state based on previous states and RTGs only.
The overall architecture of AFDT is illustrated in Fig.\ref{fig:afdt}. 
AFDT takes $K$ steps of input, consisting of $2K$ tokens, where each step contains a state and an RTG.
Similar to DT, states and RTGs are first mapped to token embedding via separate single-layer state and return-to-go encoders $\text{Embed}_s$ and $\text{Embed}_R$.
The positional embedding mapped from time steps $t$ by a single-layer temporal encoder $\text{Embed}_t$ is then added to the token embedding to include temporal information, followed by layer normalization.
These token embeddings are then processed by a GPT model \cite{radford2018improving}.
The next states are generated from the processed RTG tokens through a single-layer decoder $\text{Pred}_s$.
Note that we don't predict the next state $s_{t+1}$ directly, but rather predict the state change $\Delta s_{t+1} = s_{t+1} - s_t$ first and add it back to $s_t$ to obtain $s_{t+1}$. 
This is a common practice in motion forecasting (e.g., \citet{mohamed2020social, salzmann2020trajectron++}) to improve the prediction accuracy and has been observed to improve the performance of our model in experiments.
The algorithm of AFDT is listed in Algo.\ref{alg:afdt}.


\paragraph{Training}
At each training step, a batch of trajectories truncated to length $K$ is randomly sampled from the dataset. Each trajectory contains states and precomputed RTGs, represented as $\tau = (s_{t-K+1}, \hat R_{t-K+1}, ...,  s_{t}, \hat R_{t})$.
The model is trained autoregressively with L1 loss to predict the next state from the processed RTG token at each time step, using a causal mask to mask out future information.

\begin{algorithm}[tb]
   \caption{\Mname}
   \label{alg:afdt}
\begin{algorithmic}
   \STATE {\bfseries Input:} states $s$, returns-to-go $\hat R$, time steps $t$
   \STATE \textcolor{cyan}{\textit{ \# get positional embedding for each time step}}
   \STATE $f_t$ = $\text{Embed}_t(t)$
   \STATE \textcolor{cyan}{\textit{ \# compute the state and return-to-go embeddings}}
   \STATE $f_s,~f_{\hat R}$ = $\text{Embed}_s(s) + f_t,~\text{Embed}_R(\hat R) + f_t$
   \STATE \textcolor{cyan}{\textit{ \# send to transformer in the order ($s_0, \hat R_0, s_1, \hat R_1, ...$)}}
   \STATE $f_{output}$ = Transformer(stack($f_s$, $f_{\hat R}$))
   \STATE \textcolor{cyan}{\textit{\# predict the state change}}
   \STATE $\Delta s$ = $\text{Pred}_s$(unstack($f_{output}$.states)) 
   \STATE {\bfseries Output:} $\Delta s$ + s 
\end{algorithmic}
\end{algorithm}

\subsection{Guided Soft Actor-Critic}
\label{sec:gsac}

Now we illustrate how to use the AFDT model to benefit the learning of Soft Actor-Critic (SAC) \cite{haarnoja2018soft, haarnoja2018soft_2}. 
As AFDT can conduct planning in the state space and infer the subsequent states that lead to a high future return, our idea is to guide the agent to follow AFDT's planning.
Our method, named \emph{Guided SAC}, contains the following three main procedures.

\paragraph{Guiding Reward} We first design an intrinsic reward $r_{g,t}$, named \emph{guiding reward}, which is the discrepancy between the planned state $\widetilde s_{t+1}$
inferred by AFDT and the actual state $s_{t+1} \sim \mathrm P(\cdot|s_t, a_t)$ achieved by the agent: 
\begin{equation}
\label{eq:guidance_reward}
    r_{g,t} = -\| \frac{1}{\sigma_{\mathcal D}} \odot (\widetilde s_{t+1} - s_{t+1}) \|_2
\end{equation}
where $\sigma_{\mathcal D}$ is the standard deviation of the states over the entire offline dataset ${\mathcal D}$ and is used to normalize the different-scale state values on different dimensions.
The process to compute the guiding reward with the AFDT model is summarized in Algo.\ref{alg:guidance_reward}.

\begin{algorithm}[tb]
   \caption{Compute Guiding Reward}
   \label{alg:guidance_reward}
\begin{algorithmic}
   \STATE {\bfseries Input:} states $s_{1:t}$, return-to-go $\hat R_{1:t}$, policy $\pi$, state standard deviation $\sigma_s$, environment $env$, AFDT with context length $K$
   \REPEAT
   \STATE \textcolor{cyan}{\textit{ \# get AFDT's prediction of the next state}}
   \STATE $\widetilde s_{t+1}$ = AFDT($s_{t-K+1:t}$, $\hat R_{t-K+1, t}$)
   \STATE \textcolor{cyan}{\textit{ \# apply the policy in the environment for one step}}
   \STATE $a_t$ = $\pi(s_t)$
   \STATE $s_{t+1}$, $r_e$ = $env$.step($a_t$)
   \STATE \textcolor{cyan}{\textit{ \# compute current guiding reward using Eq.\ref{eq:guidance_reward}}}
   \STATE $r_{g}$= $-\| \frac{1}{\sigma_s} \odot (\widetilde s_{t+1} - s_{t+1}) \|_2$
   \STATE \textcolor{cyan}{\textit{ \# update return-to-go (same as DT) and time step}}
   \STATE $\hat R_{t+1}$ = $\hat R_{t}$ - $r_e$
   \STATE $t = t + 1$
   \UNTIL{Episode is finished}
\end{algorithmic}
\end{algorithm}

\begin{figure*}[t]
\centering
 \begin{subfigure}[b]{0.32\textwidth}
    \includegraphics[width=\textwidth]{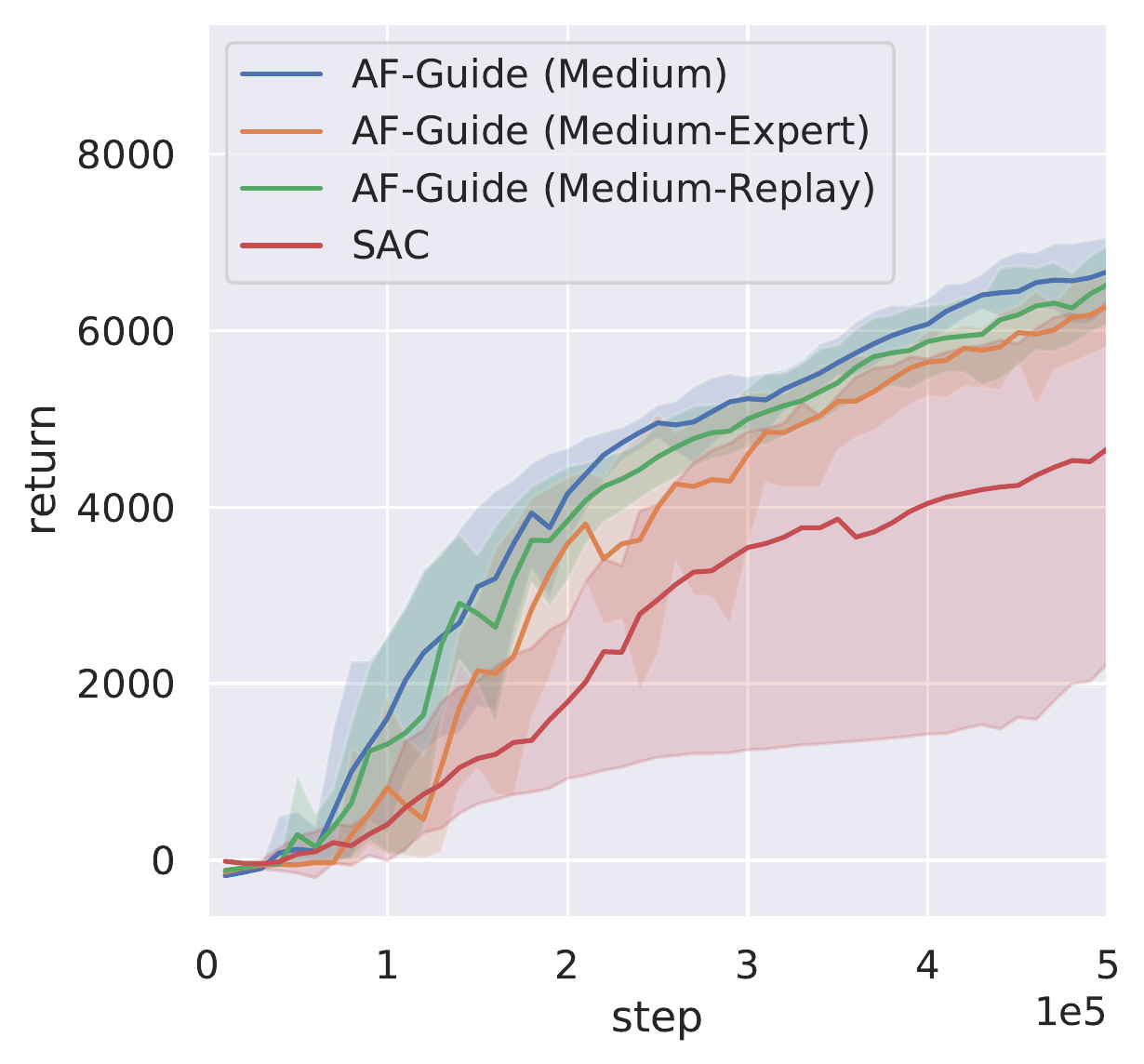}
    \caption{Halfcheetah}
\end{subfigure}
\begin{subfigure}[b]{0.32\textwidth}
    \includegraphics[width=\textwidth]{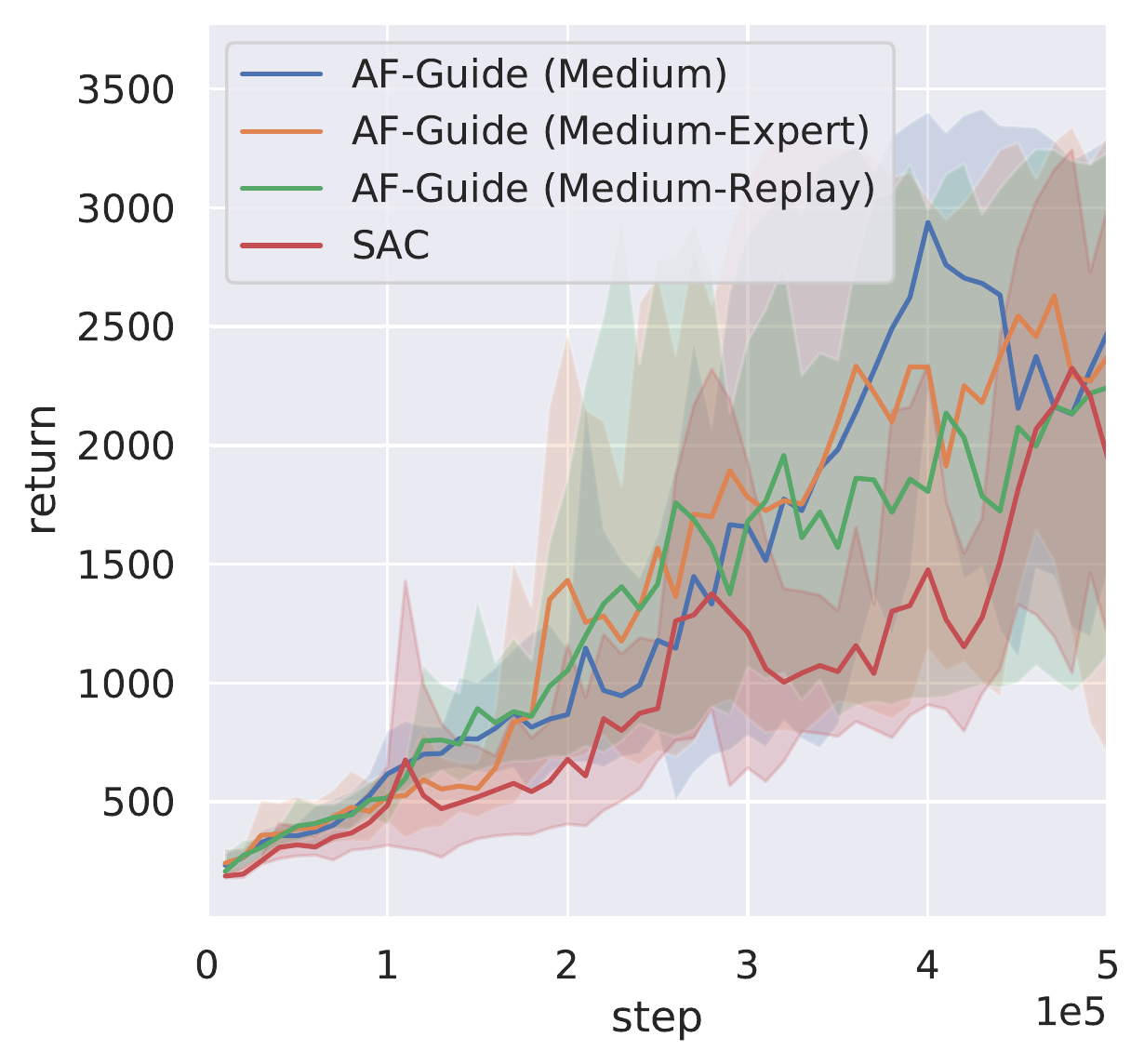}
    \caption{Hopper}
\end{subfigure}
\begin{subfigure}[b]{0.32\textwidth}
    \includegraphics[width=\textwidth]{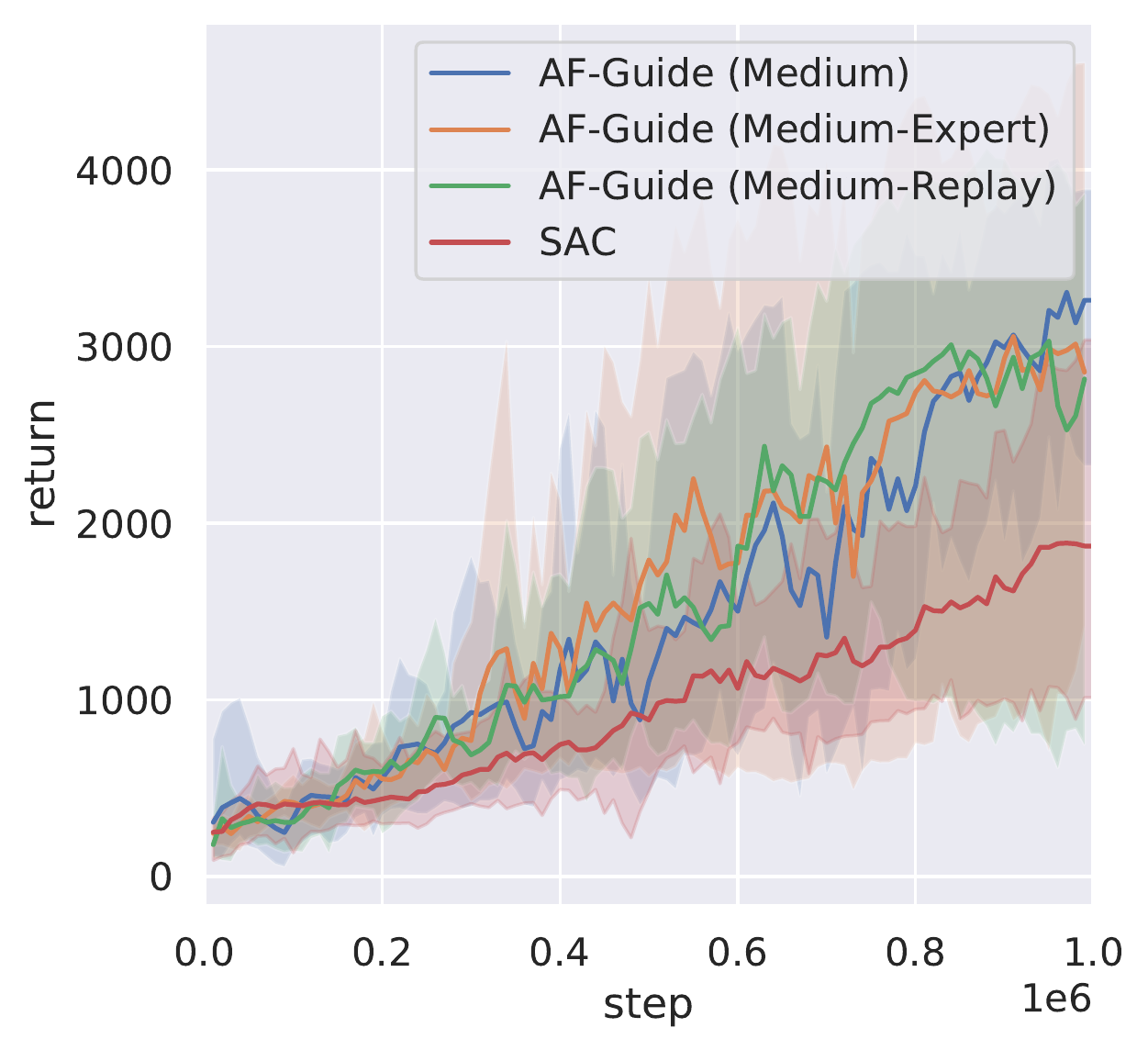}
    \caption{Walker2d}
\end{subfigure}
\begin{subfigure}[b]{0.32\textwidth}
    \includegraphics[width=\textwidth]{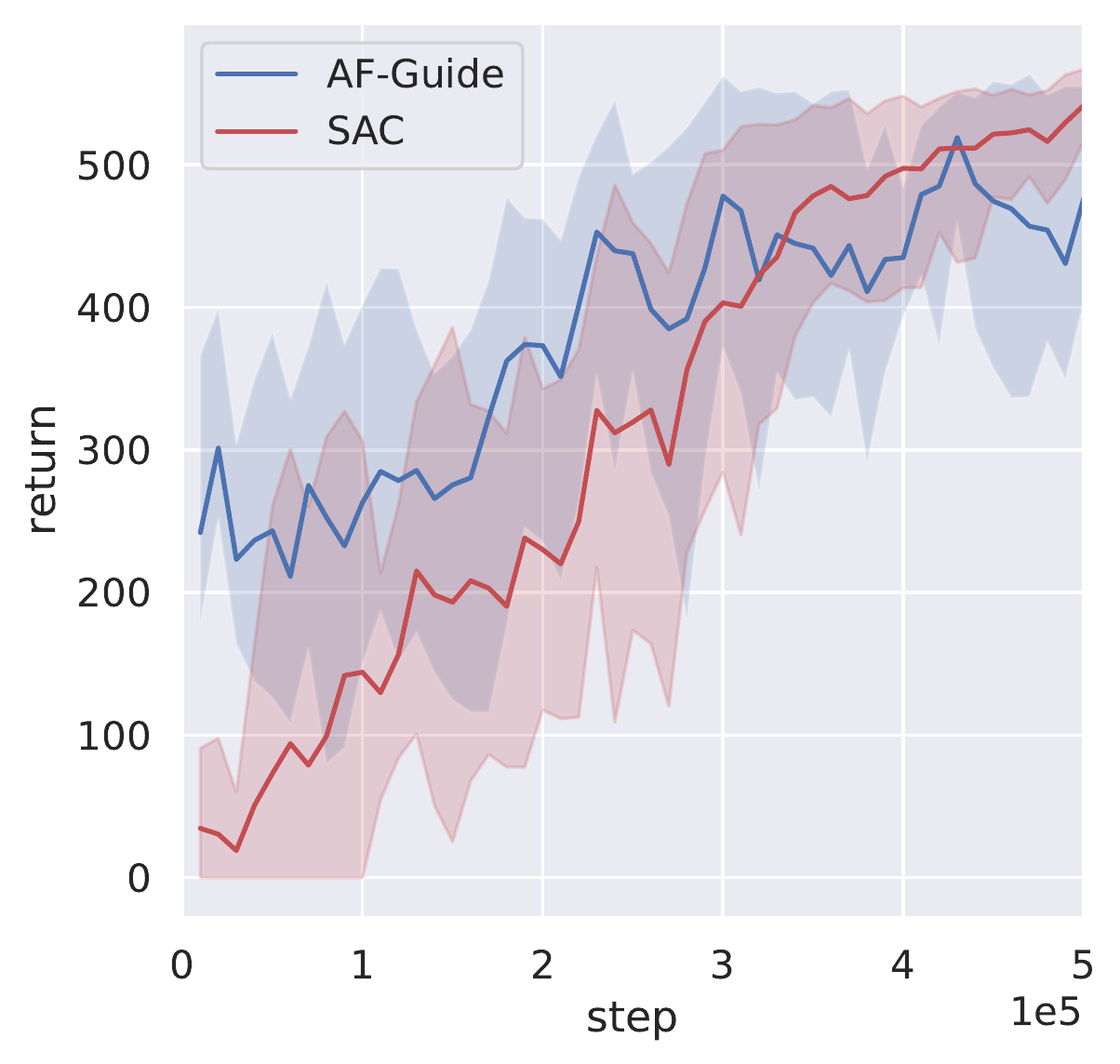}
    \caption{Maze2d-Medium}
\end{subfigure}
\begin{subfigure}[b]{0.32\textwidth}
    \includegraphics[width=\textwidth]{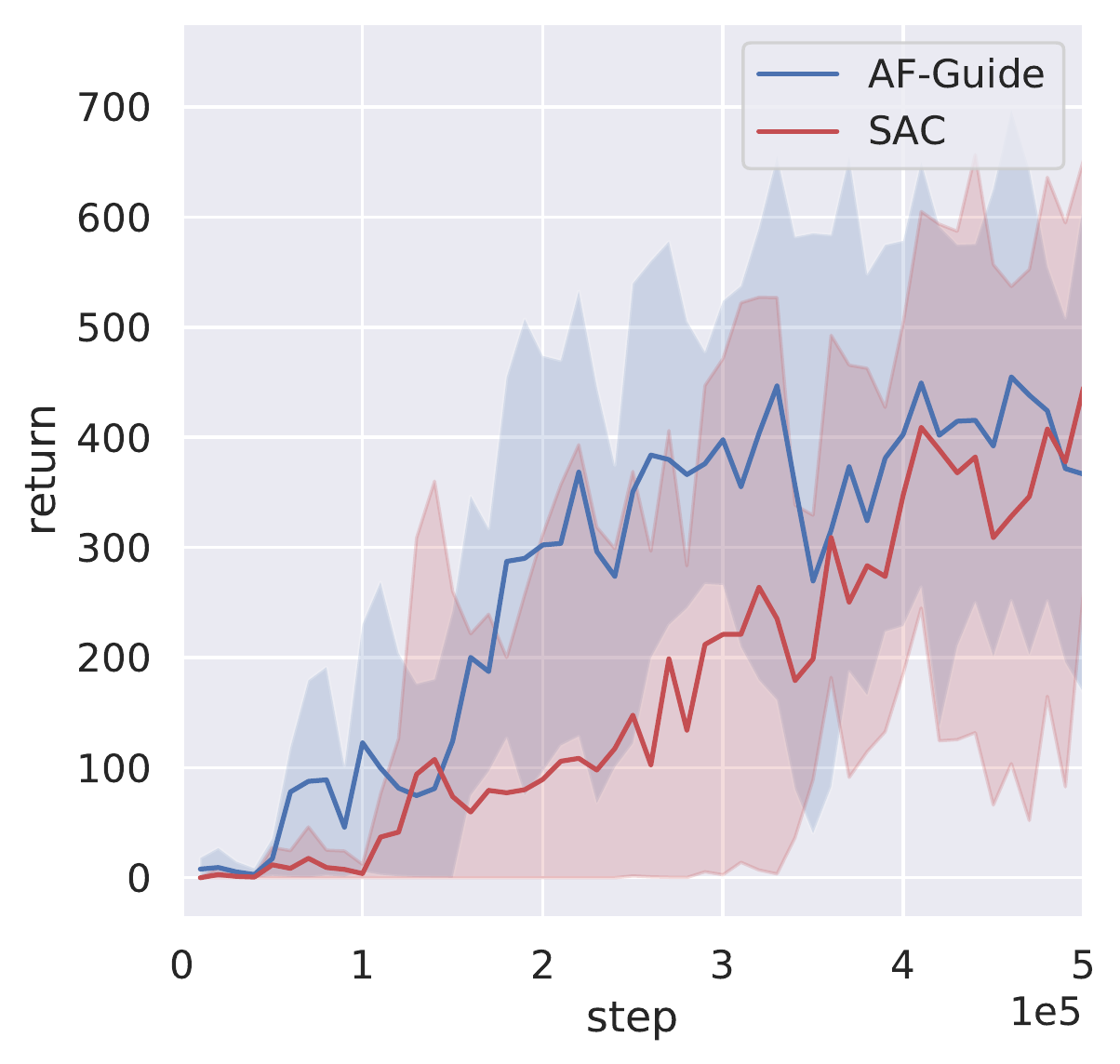}
    \caption{Maze2d-Large}
\end{subfigure}
\begin{subfigure}[b]{0.32\textwidth}
    \includegraphics[width=\textwidth]{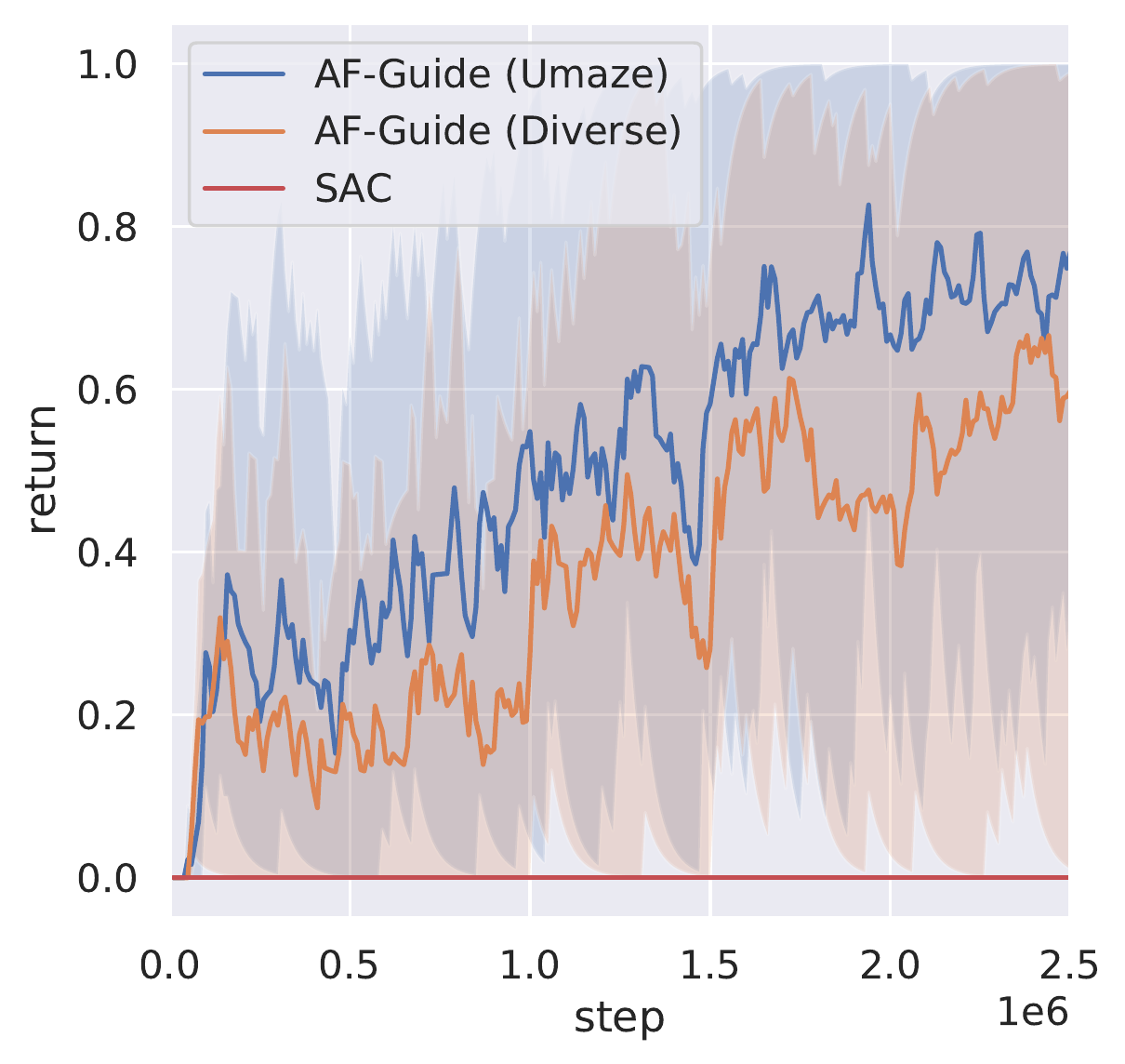}
    \caption{Antmaze-Umaze}
\end{subfigure}
\caption{Experimental results of our methods. Utilizing the knowledge learned from the action-free offline dataset, \Mabbr~outperforms SAC in all evaluated locomotion and ball maze environments in terms of learning speed. 
Furthermore, while SAC struggles to complete the task of Antmaze-Umaze due to the challenging exploration, \Mabbr~successfully solves it, owing to the guidance signals provided by AFDT.}
\label{fig:exp}
\end{figure*}

\paragraph{Guiding Q Function}
We then use the guiding reward $r_g$ to learn the Q function.
A common practice is to combine the intrinsic reward and the environment reward by $r_t=r_{e,t}+\beta 
 r_{g,t}$ with a coefficient $\beta$, and use a single Q network $Q$ to approximate the long-term future return \cite{schmidhuber1990making, schmidhuber1991possibility, houthooft2016vime, pathak2017curiosity, tao2020novelty}. 
However, this is not the case for the guiding reward, where \emph{the current action should only be responsible for the next immediate result rather than all the future results}. Assume a robot gets stuck at step $t+1$ due to a bad action $a_t$ at step $t$. 
A good AFDT will give the robot a low guidance reward at step $t$ and predict a static future, resulting in high future guidance rewards for getting the robot stuck in the same state.
More generaly, as AFDT replans the target states at every timestep,
an agent missing the planned state $\tilde s_{t}$ due to a bad action $a_{t-1}$ can still reach the replanned state $\tilde s_{t+1}$ at the next step and receive a high guiding reward $r_{g, t}$, which is not desirable.
Hence, the action $a_t$ should not be rewarded by $r_{g, t+1}$ as it didn't reach the original plan $\tilde s_t$.
Therefore, to prevent the guiding reward from misleading the agent, it is more reasonable to discard the future return for the Q value calculation of the current action. 


Due to the reason above, we set up an additional independent Guiding Q function $Q_g$ which is optimized in the same way as the original Q function $Q_e$ (see Eq.\ref{eq_Q_SAC}), 
but the target Q value only involve the immediate reward $r_{g,t}$ without future information, which is computed as follows:
\begin{equation}
    Q_{g,t}^{\rm target} = r_{g,t}
\end{equation}
Compared to Eq.\eqref{eq_Q_target_SAC}, here the Q target of the current action is removed from the future information by setting the discount factor $\gamma$ to zero.
Our ablation study in Sec.\ref{sec: ablation} demonstrates that the Guiding Q function is crucial to effective guidance.

\paragraph{Combined Q function}
We finally replace the Q function $Q_e$ in Eq.\eqref{eq_sac_policy} with the following combined Q function to guide the policy learning:
\begin{equation}
    Q(s_t, a_t) = Q_e(s_t, a_t) + \beta Q_g(s_t, a_t)
\label{eq: q function}
\end{equation}
where $\beta$ is the coefficient. 
Note that when $\beta=0$, Guided SAC degenerates to a standard SAC trained using environment rewards $r_e$ and the corresponding Q function $Q_e$ only.

\section{Experiments}

\begin{figure*}[t]
\centering
 \begin{subfigure}[b]{0.31\textwidth}
    \includegraphics[width=\textwidth]{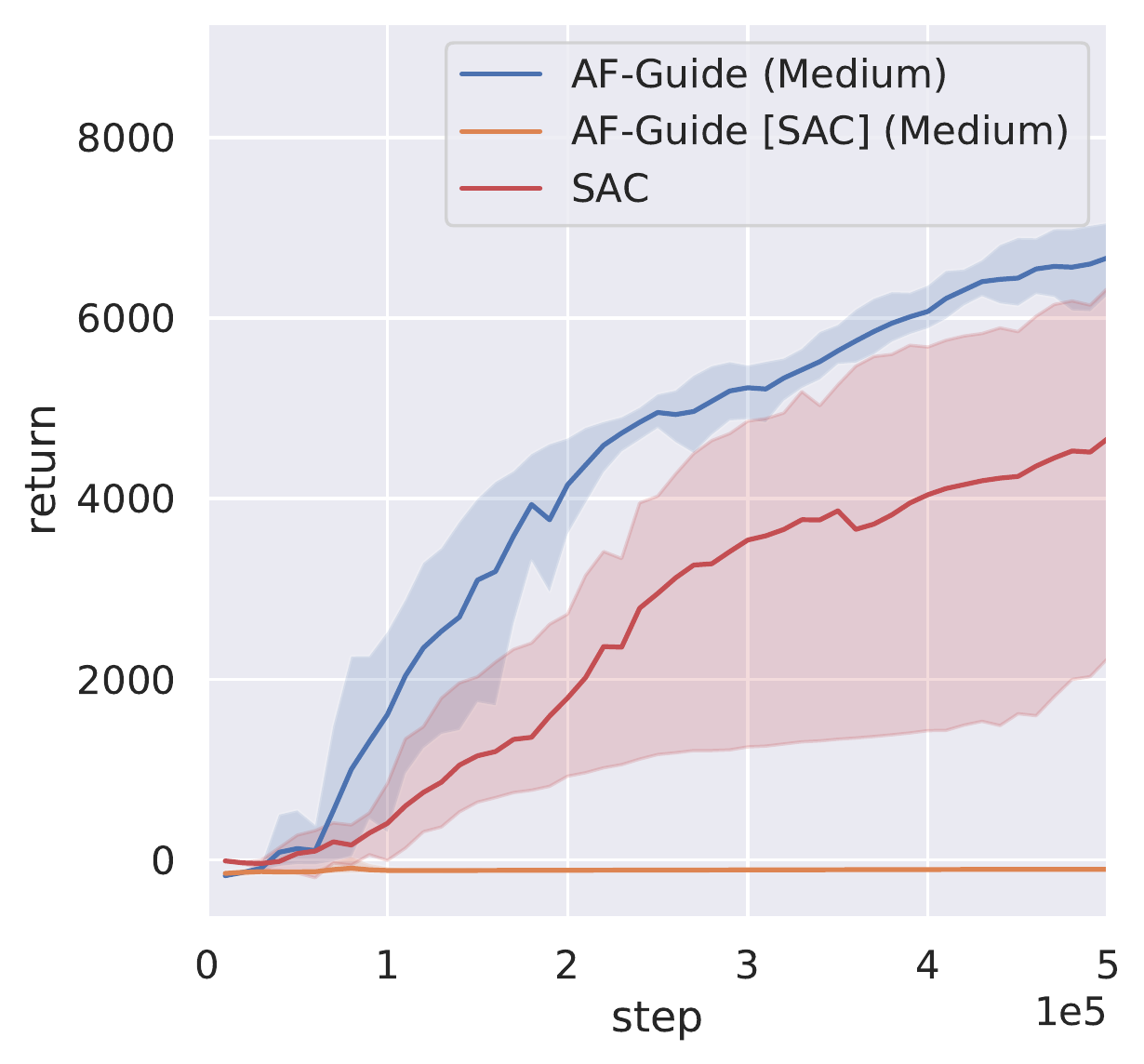}
    \caption{Halfcheetah}
\end{subfigure}
\begin{subfigure}[b]{0.32\textwidth}
    \includegraphics[width=\textwidth]{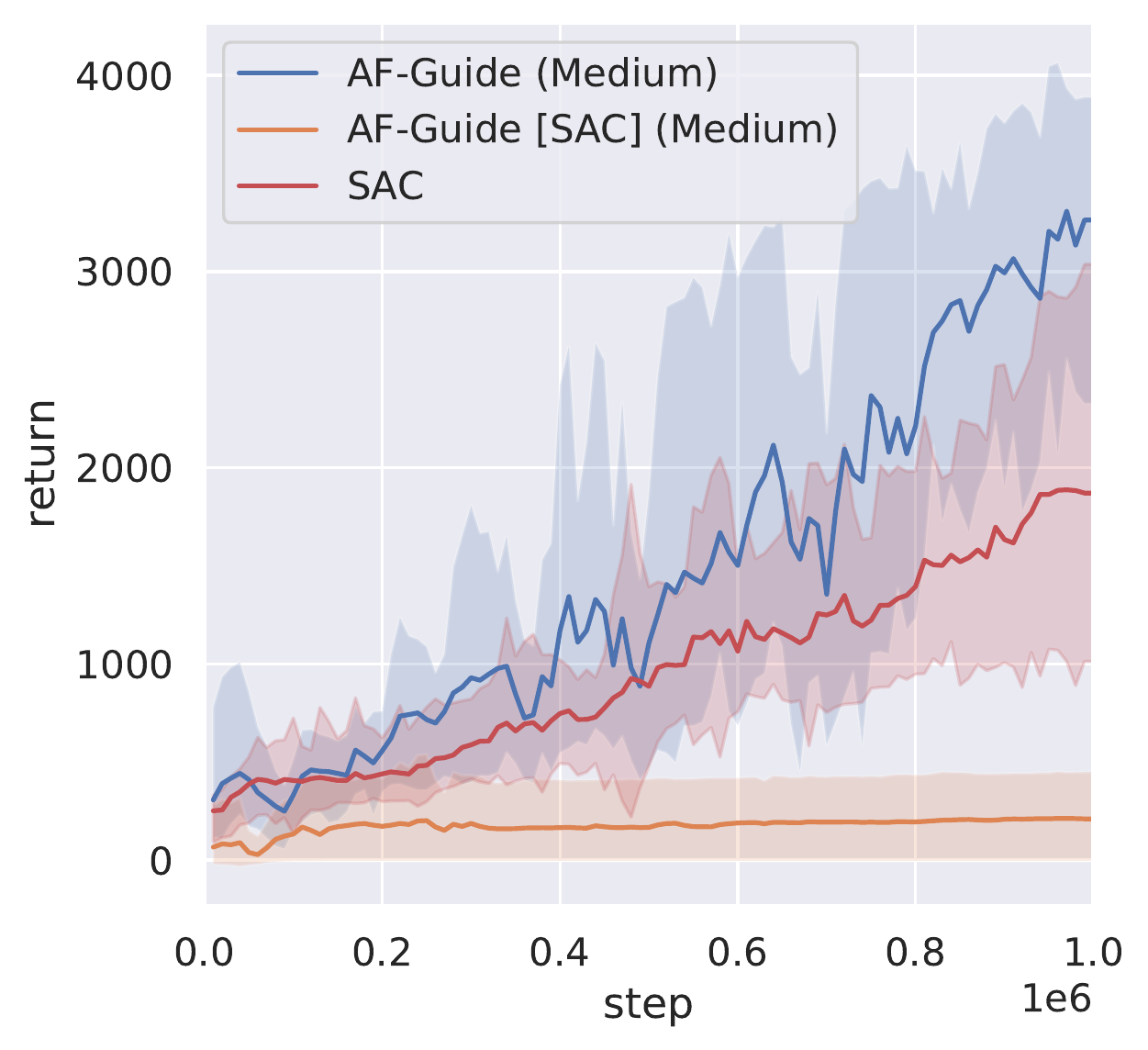}
    \caption{Walker2d}
\end{subfigure}
\begin{subfigure}[b]{0.31\textwidth}
    \includegraphics[width=\textwidth]{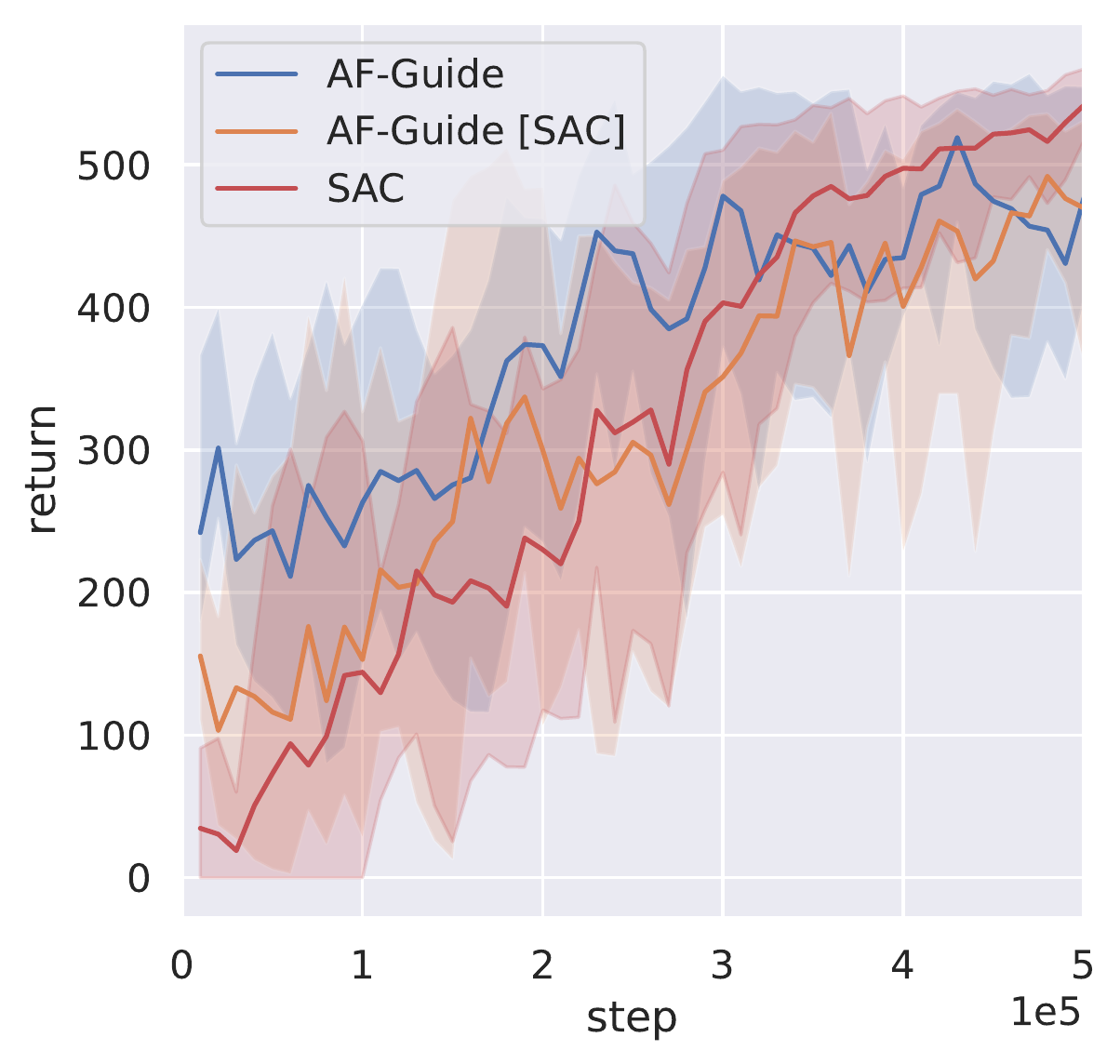}
    \caption{Maze2d-Medium}
\end{subfigure}
\caption{Ablation study on the usage of guiding reward $r_g$. `\Mabbr~[SAC]' denotes the variant adding guiding reward to the environment reward and training with SAC. 
The results show that \Mabbr~[SAC] performs similarly to SAC in Maze2d-Medium, but does not work in Halfcheetah and Walker2d, which indicates that simply adding the guiding reward is detrimental to the policy training and verifies the effectiveness of our Guided SAC design.}
\label{fig:ablation_gsac}
\end{figure*}

In this section, we demonstrate the effectiveness of our approach \Mabbr~for utilizing action-free offline reinforcement learning datasets in online reinforcement learning through experimental evaluation.
Furthermore, we provide evidence for the validity of our design choices for the two components of \Mabbr, Action-Free Decision Transformer, and Guided SAC, through three ablation studies.

\paragraph{Action-Free D4RL Benchmark}
To evaluate methods on AFP-RL,
we build on top of the widely-used offline reinforcement learning benchmark, D4RL \cite{fu2020d4rl}, and adapt it to the action-free reinforcement learning setting. We denote the introduced benchmark as Action-Free D4RL.
The original D4RL benchmark provides offline datasets collected using various strategies across different environments. 
These episodes in the original D4RL datasets include state, action, and reward sequences. 
To create our action-free offline RL datasets, we remove the action labels from the original datasets. 
We evaluate six environments, including three locomotion tasks (Hopper, Halfcheetah, Walker2d), two ball maze environments (Maze2d-Medium, Maze2d-Large), and one robot ant maze environment (Antmaze-Umaze). 
For each locomotion task, we test our method on three different datasets: Medium, Medium-Replay, and Medium-Expert. 
For the environment Antmaze-Umaze, we test on two datasets: Antmaze-Umaze and Antmaze-Umaze-Diverse. 
There is only one dataset for each ball maze environment, where the ball navigates to random goal locations. 
Details of the datasets can be found in Appx.\ref{appx: benchmark}.

\paragraph{Implementation Details}
The training of \Mabbr~contains two stages: an offline stage training AFDT using the offline dataset and an online stage training Guided SAC in the environment.
For the architecture and the training of AFDT, we follow the default hyperparameters used in DT paper \cite{chen2021decision}. 
The context length $K$ is set to 20. The batch size for AFDT training is 64 and the learning rate is 1e-4 with AdamW optimizer.
In the online training stage, we set the return-to-go $\hat R$ to 6000, 3600, and 5000 for Halfcheetah, Hopper, and Walker2d, respectively, the same as the values used in the original DT paper. 
The robot ant maze environment and the ball maze environments are not used in the original DT paper.
We set $\hat R$ to 1 and 5000, separately.
For the hyperparameters of Guided SAC, we follow the default hyperparameters of SAC in the widely used Stable Baseline 3 \cite{stable-baselines3} implementation. 
The training batch size is 256 and the learning rate is 3e-4 with Adam optimizer.
The discount factor for the environment reward is 0.99. 
The coefficient of the Guided Q function $\beta$ in Eq.\ref{eq: q function} is set to 3.
More details of the hyperparameters can be found in Appx.\ref{appx: hyperparameters}.

\subsection{Main Experiments}

\begin{figure*}[t]
\centering
 \begin{subfigure}[b]{0.32\textwidth}
    \includegraphics[width=\textwidth]{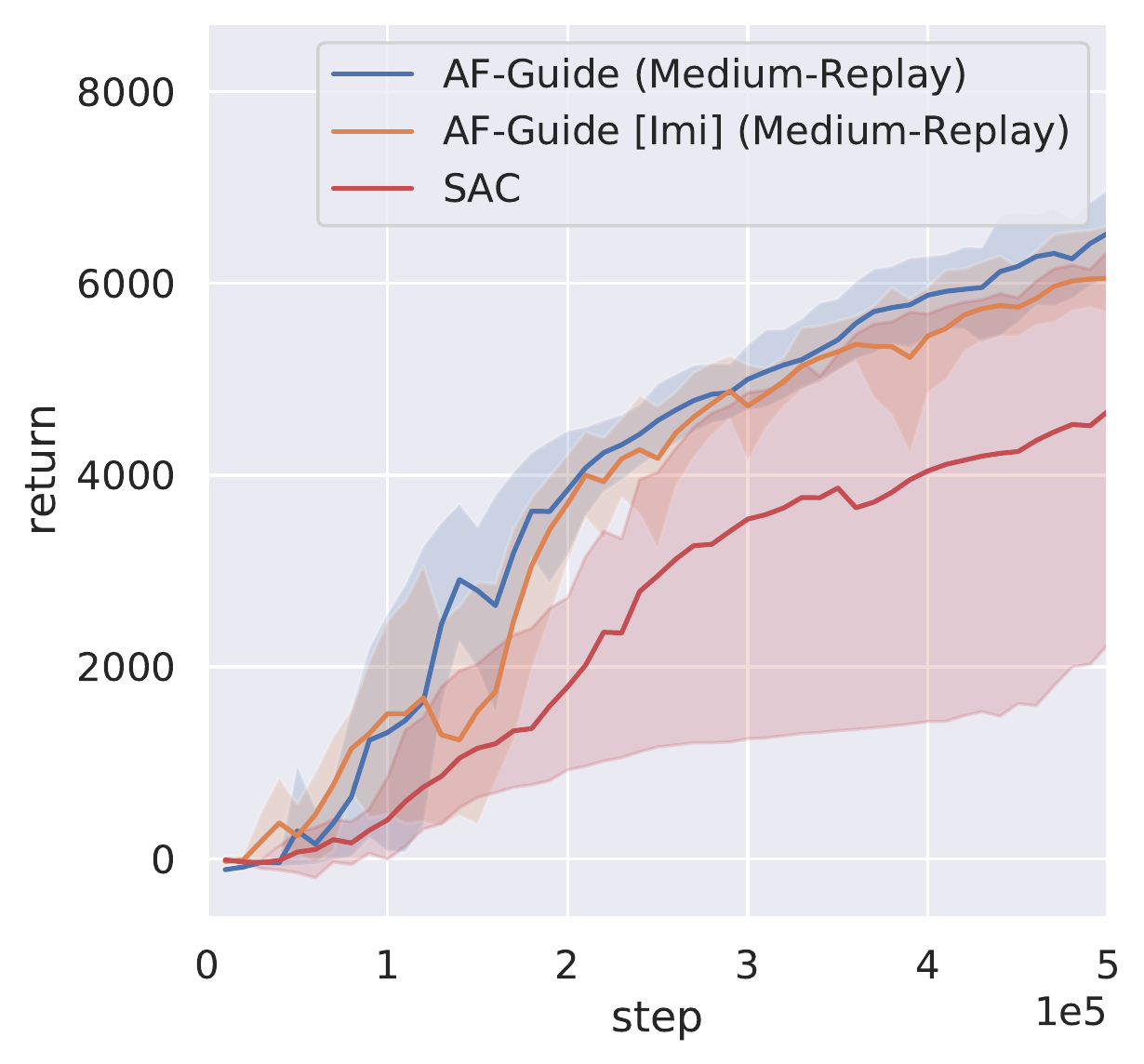}
    \caption{Halfcheetah}
\end{subfigure}
\begin{subfigure}[b]{0.32\textwidth}
    \includegraphics[width=\textwidth]{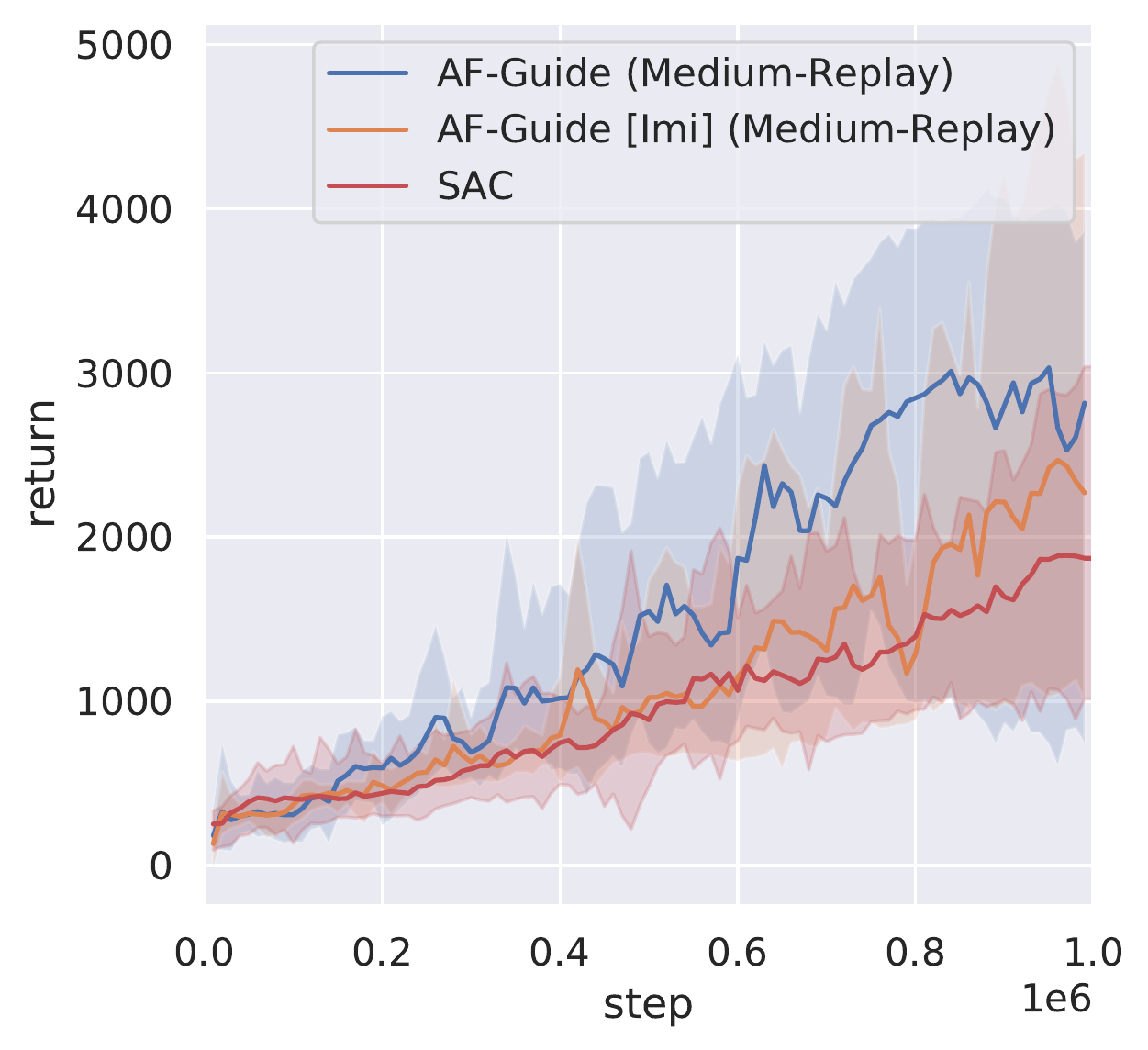}
    \caption{Walker2d}
\end{subfigure}
\begin{subfigure}[b]{0.31\textwidth}
    \includegraphics[width=\textwidth]{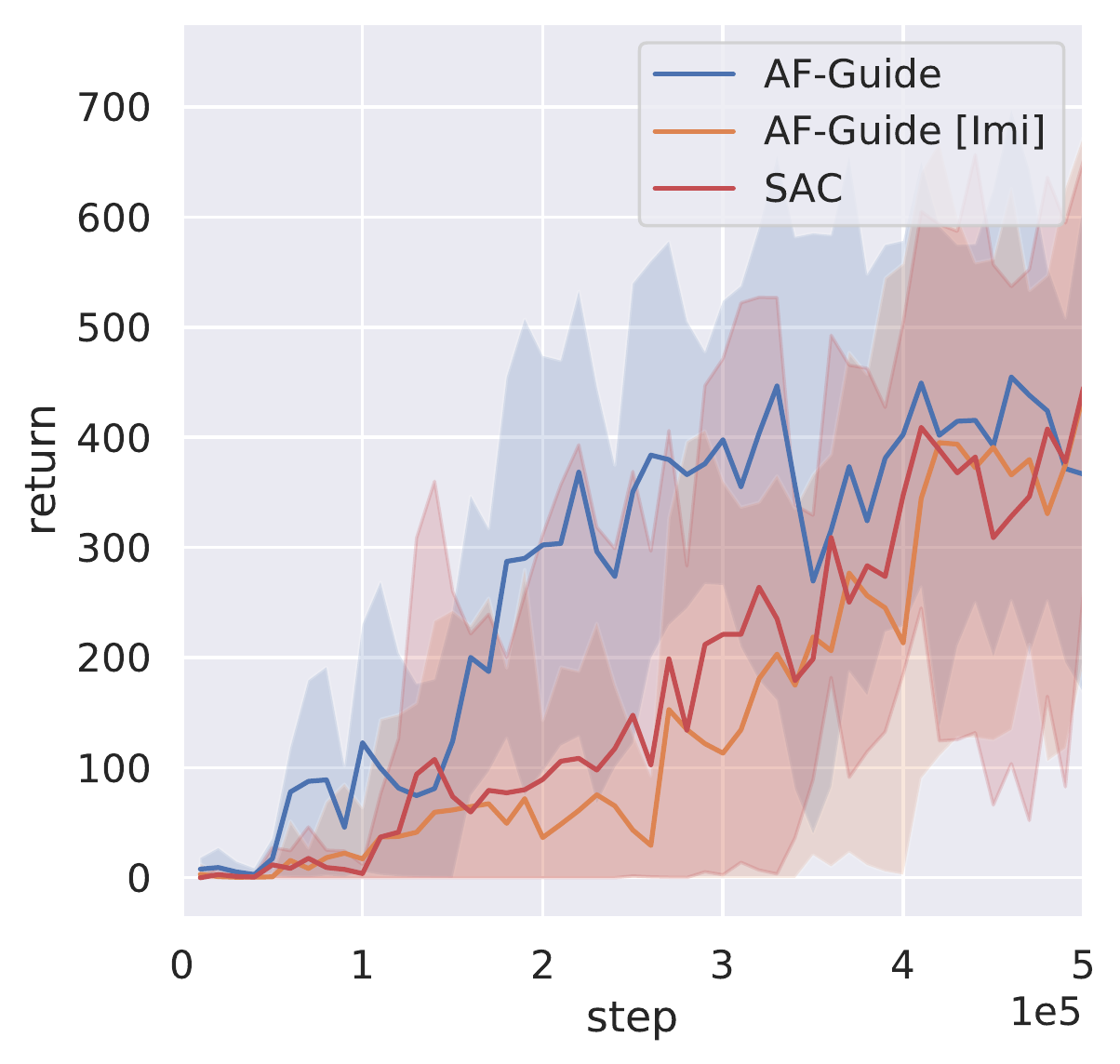}
    \caption{Maze2d-Large}
\end{subfigure}
\caption{Ablation study on the effectiveness of Action-Free Decision Transformer (AFDT). We train a variant of AFDT by regressing the behavior policy trajectories, and use this variant to guide the online training, referred to as \Mabbr~[Imi]. 
Compared to \Mabbr, \Mabbr~[Imi] performs worse in Walker2d and Maze2d-Large. 
This suggests that AFDT can infer the next states better than those collected by the behavior policy.}
\label{fig:ablation_bc}
\end{figure*}

\paragraph{Results Analysis}
Experimental results are presented in Figure \ref{fig:exp}. We run each experiment four times, using different random seeds, and report the average and the standard deviation band. Our method \Mabbr~, using knowledge learned from the action-free offline dataset, outperforms SAC in all the evaluated environments. 
In the tasks of Halfcheetah and Walker2d, \Mabbr~shows a significant advantage in learning speed compared to SAC across all the three datasets. 
In Halfcheetah, \Mabbr~demonstrates a significant improvement of 50\% at 500k steps, with an achieved performance of 6000 compared to the 4000 achieved by SAC alone.
Similarly, in Walker2d, \Mabbr~improves the performance by 50\% at 1M steps, from 2000 to 3000. 
Additionally, we observe that different offline datasets do not result in significant performance differences. 
In the tasks of Hopper, Maze2d-Medium, and Maze2d-Large, while both \Mabbr~and SAC reach similar performance at 500k steps, \Mabbr~converges faster. 
In the task of Antmaze-Umaze, SAC is unable to complete it in 1M steps, whereas \Mabbr~has an 80\% success rate when pretrained in the dataset Antmaze-Umaze and a 60\% success rate in Antmaze-Umaze-Diverse. 
This is likely due to the exploration challenge faced by SAC. 
The robot ant in Antmaze-Umaze has 4 legs with a total of 8 joints and only receives rewards when reaching the target location, resulting in large state/action spaces and sparse reward signals. 
As a result, the agent trained by SAC does not receive any rewards during exploration and does not know how to move. 
In contrast, our guiding reward learned from the action-free offline dataset provides dense learning signals that guide the agent's motion towards the target. 
Therefore, agents trained by \Mabbr~can successfully solve the maze in this task.

\subsection{Ablation Study}
\label{sec: ablation}

\paragraph{Do we really need Guided SAC?}
In this ablation study, we investigate whether our Guided SAC with an additional Q function is necessary to process the guiding reward $r_g$, or if it can be simply added to the environment reward and processed by SAC, referred to as `\Mabbr~[SAC]'. 
This study is conducted in the locomotion environments of Halfcheetah and Walker2d using the Medium dataset and the maze environment of Maze2d-Medium. 
The results, shown in Fig.\ref{fig:ablation_gsac}, reveal that \Mabbr~[SAC] has a similar performance as SAC in Maze2d-Medium and does not work at all in Halfcheetah and Walker2d, indicating that the guiding reward $r_g$ does not help or even hinders the training of SAC. 
In contrast, the original \Mabbr~with Guided SAC benefits from the guiding reward $r_g$ by ignoring guiding rewards in future steps and setting the corresponding discount factor to zero. 
This is in line with our explanation in Sec.\ref{sec:gsac}, where we stated that high future guiding rewards are unrelated to the action quality at the current step and therefore should be ignored in the Q function. 
Experimental results verify the effectiveness of our Guided SAC design.

\paragraph{Does AFDT plan better states than that from behavior policy?}
In this ablation study, we investigate whether the states planned by Action-Free Decision Transformer (AFDT) are superior to the original states collected by the behavior policy. 
To do this, we trained an AFDT in an `imitation' style by directly regressing future states based solely on past states without any return-to-go information in the offline dataset. 
We denote this AFDT variant as AF-Imitation.
In the online training stage, AF-Imitation plans the next state based on past states only without returns-to-go. 
We refer to the method trained with the guidance of AF-Imitation as `\Mabbr~[Imi]' and evaluate it in the locomotion environments of Halfcheetah and Walker2d using the Medium-Replay dataset and the maze environment of Maze2d-Large. 
The results, shown in Fig.\ref{fig:ablation_bc}, demonstrate that \Mabbr~[Imi] performs worse than the original version in Walker2d and Maze2d-Large and is close to the original version in Halfcheetah.
This verifies that AFDT can plan the next states that lead to higher future returns than the behavior policy when conditioned on a proper return-to-go. 
Additionally, \Mabbr~[Imi] performs better than SAC in Halfcheetah and slightly better in Walker2d. While the predicted state from \Mabbr~[Imi] may not be optimal, it can still benefit policy training in some cases.

\begin{table}[t]
\caption{Training time comparison of \Mabbr~[TT] and \Mabbr~for 500k environment steps in locomotion tasks. \Mabbr~[TT] increases the training time dramatically, due to the huge planning cost in the original Trajectory Transformer design. Experiments are done using a single A100 GPU.}
\vskip 0.15in
\begin{center}
\begin{small}
\begin{sc}
\begin{tabular}{lcc}
\toprule
Env. & \Mabbr~[TT] & \Mabbr \\
\midrule
Halfcheetah   & $\sim$14 hours  & $\sim$1 hour \\
Hopper        & $\sim$10 hours  & $\sim$1 hour \\
Walker2d      & $\sim$20 hours  & $\sim$1 hour \\
\bottomrule
\end{tabular}
\end{sc}
\end{small}
\end{center}
\label{tab: tt time}
\end{table}

\begin{figure*}[t]
\centering
 \begin{subfigure}[b]{0.32\textwidth}
    \includegraphics[width=\textwidth]{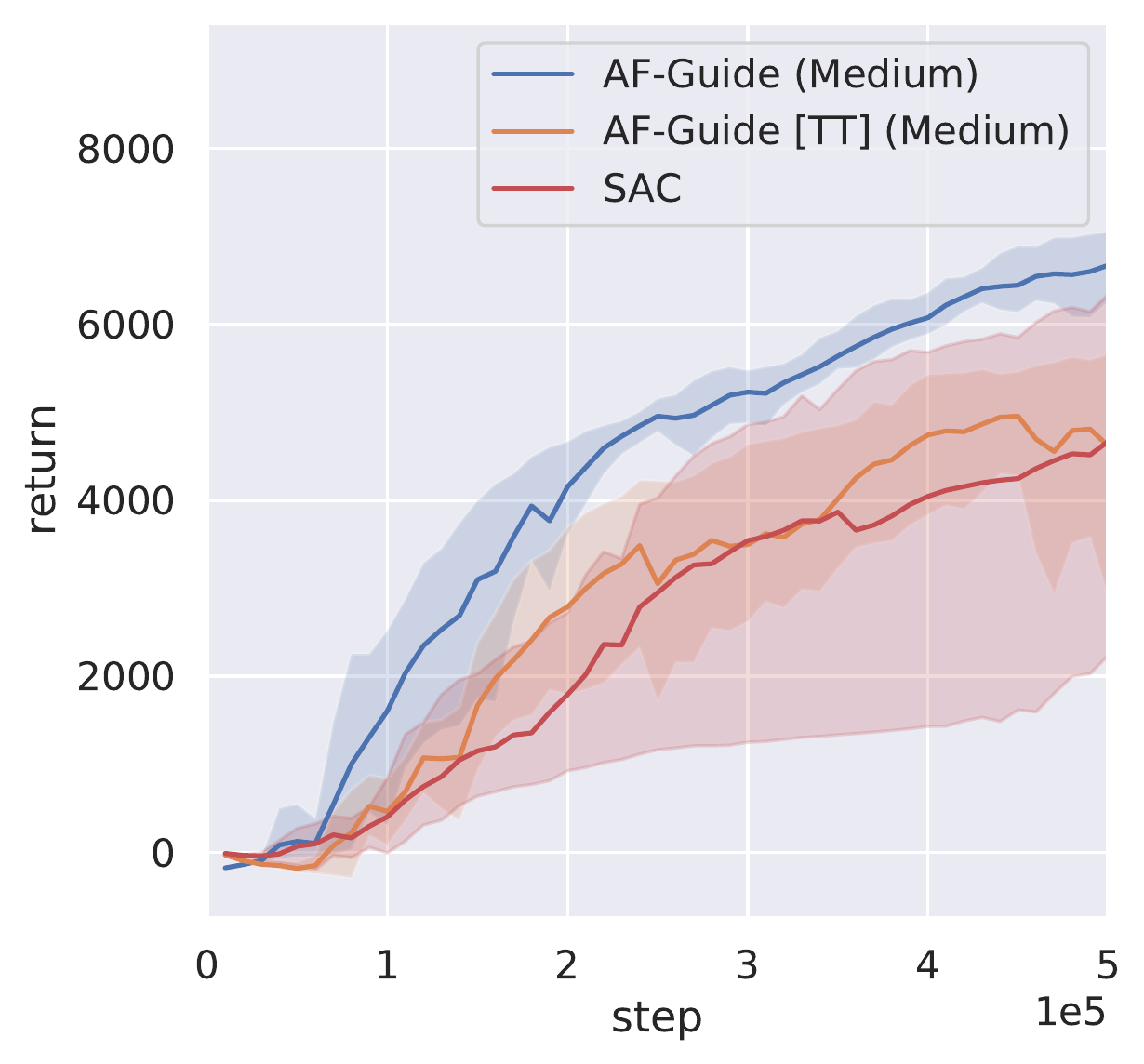}
    \caption{Halfcheetah}
\end{subfigure}
\begin{subfigure}[b]{0.32\textwidth}
    \includegraphics[width=\textwidth]{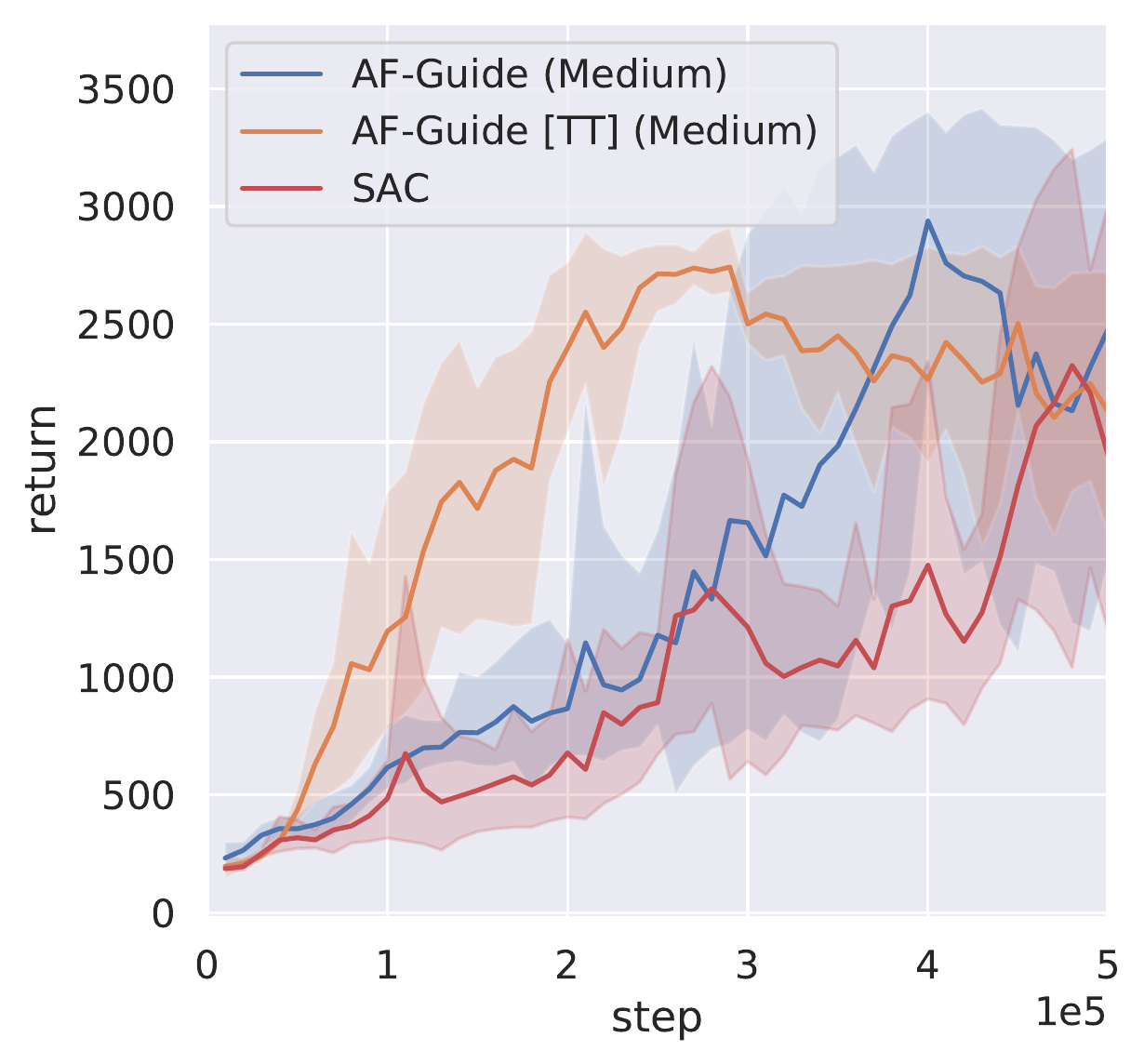}
    \caption{Hopper}
\end{subfigure}
\begin{subfigure}[b]{0.32\textwidth}
    \includegraphics[width=\textwidth]{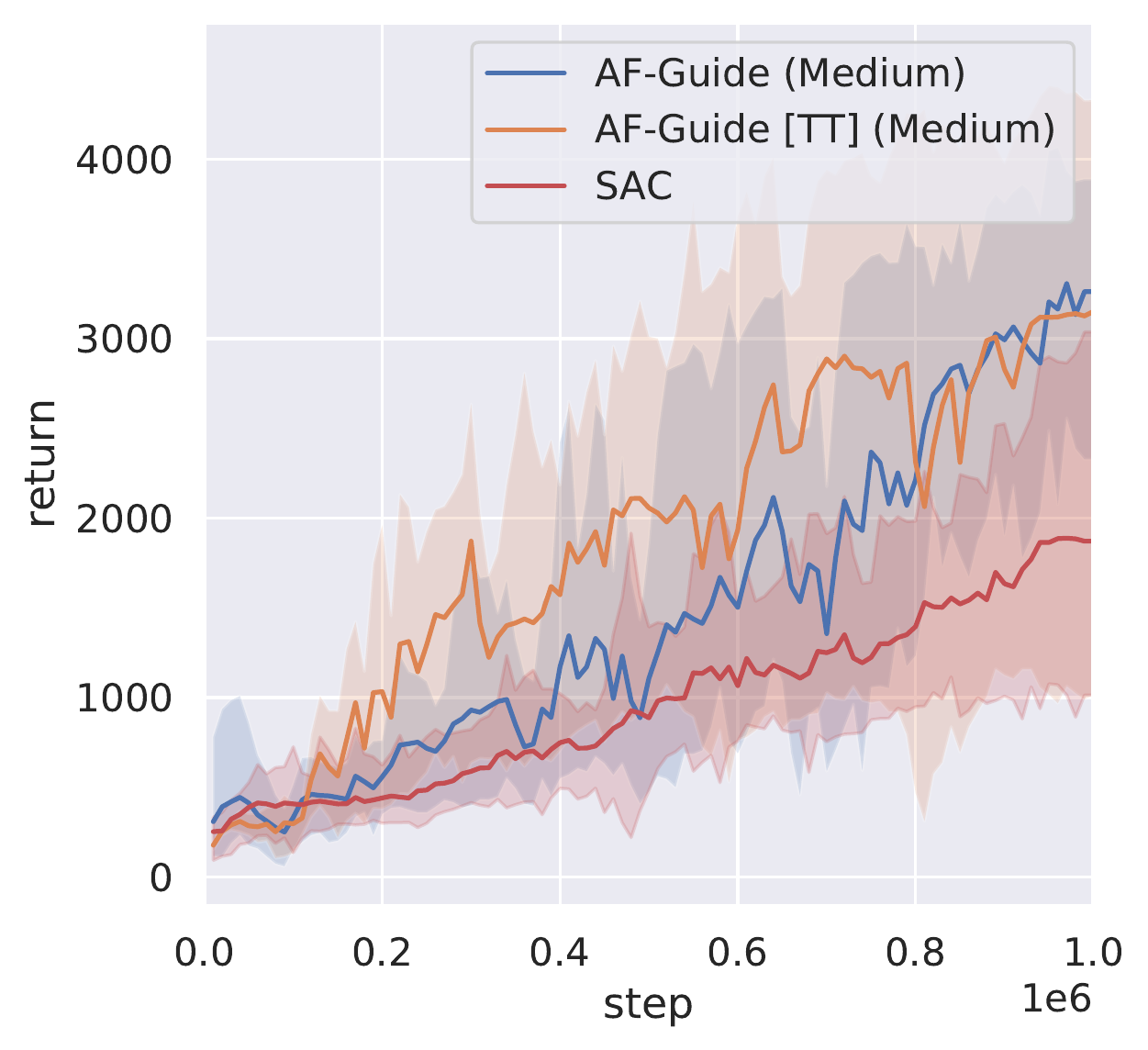}
    \caption{Walker2d}
\end{subfigure}
\caption{Ablation study on using Action-Free Trajectory Transformer to guide the training (\Mabbr~[TT]). The results showed that \Mabbr~[TT] had better performance in the Hopper and Walker2d tasks, but performed worse in the Halfcheetah task. These results suggest that our pipeline is compatible with different sequential-modeling-based offline RL methods, but the choice of method may impact performance depending on the specific task.}
\label{fig:ablation_tt}
\end{figure*}

\paragraph{Can Action-Free Trajectory Transformer replace Action-Free Decision Transformer?}
Guided SAC is built on the predictions of AFDT, our action-free variant of the Decision Transformer (DT). 
In theory, AFDT can be replaced by any other sequential-modeling-based offline RL method after removing the action information. 
In this experiment, we replaced AFDT with an action-free variant of Trajectory Transformer (TT) \cite{janner2021offline}, and evaluated its performance on locomotion tasks using the Medium dataset.
We denote this variant as \Mabbr~[TT].
Details of \Mabbr~[TT] can be found in Appx.\ref{appx:aftt}.
Compared to DT, which generates the future in a UDRL style, TT directly rollouts future predictions via beam search and selects the predictions with the highest predicted future returns. 
Additionally, TT discretizes the state and action spaces dimension-by-dimension for improved prediction accuracy. 
Experimental results are shown in Fig.\ref{fig:ablation_tt}. 
\Mabbr~[TT] performs worse than \Mabbr~in Halfcheetah, but it showed a clear advantage over \Mabbr~in Hopper and Walker2d. 
This advantage of \Mabbr~[TT] in Hopper and Walker2d may be due to the better prediction quality from the dimension-wise discretization. 
However, the training time of \Mabbr~[TT] is much longer than \Mabbr. 
As TT discretizes the state dimension-by-dimension, predicting one step of the state requires multiple forward passes of the transformer model, and to pick the best prediction, TT needs to generate multiple rollouts via beam-search. 
This dramatically increases the computational cost of future reasoning and slows down the training.
A brief comparison of the training time between \Mabbr~and \Mabbr~[TT] is shown in Tab.\ref{tab: tt time}. 
\Mabbr~[TT] is at least 10 times slower than \Mabbr~in our experiments. 
The more dimensions the state space has, the slower \Mabbr~[TT] is.
Therefore, we select Decision Transformer in our method. 
The experiments show that our pipeline is compatible with different sequential-modeling-based offline RL methods.

\subsection{Limitations}
As an attempt to utilize action-free offline datasets for improved online learning, our method has some limitations in its current form. 
Firstly, our current design of \Mabbr~is based on Decision Transformer following UDRL framework \cite{schmidhuber2019reinforcement}. 
Therefore, our planning ability is also limited by Decision Transformer's design and UDRL's drawback:
\citet{vstrupl2022upside} show that UDRL may diverge from the optimal policy in an episodic setting with stochastic environments.
Note that the framework of our method \Mabbr~is agnostic to the sequential planning model as we show in the ablation study with Trajectory Transformer based on beam-search instead of UDRL in Sec.\ref{sec: ablation}.
Secondly, our current guiding reward is based on L2 distance, which may not be optimal in some state spaces where L2 distance doesn't represent the state similarities well, such as images
We believe that combining \Mabbr~with more semantically meaningful similarity metrics can extend its applications in the future for vision, language, and other multimodal problems.
We leave this for future research to explore.


\section{Conclusion}

In this paper, we explore the potential of utilizing action-free offline datasets to guide online reinforcement learning, and denote this setting by Reinforcement Learning with Action-Free Offline Pretraining (AFP-RL). 
We propose \Mname~(\Mabbr), a method that learns to plan the target state from offline datasets to guide the online learning of an SAC agent. 
Our experimental results demonstrate that \Mabbr~has better sample efficiency than SAC in various locomotion and maze environments, highlighting the benefits of incorporating action-free offline datasets. We hope our work may encourage further research in other areas where action-free offline pretraining can be an effective learning approach. These applications may include 
combining video prediction models with semantically meaningful similarity metrics to build guidance rewards learned from large-scale Internet data.



\bibliography{example_paper}
\bibliographystyle{icml2023}

\newpage
\appendix
\onecolumn
\section{Datasets}
\label{appx: benchmark}
For locomotion environments, 
the Medium dataset is collected using a policy trained to approximately 1/3 the performance of an expert. 
Medium-Replay uses the training replay buffer of the 'Medium' policy. 
The Medium-Expert dataset contains 50\% of data from Medium and the remaining data is collected by an expert policy. 
For the ant robot maze environment,
in the Antmaze-Umaze dataset, the robot ant always goes from a fixed start position to a fixed target location, while in Antmaze-Umaze-Diverse, the robot ant goes to random target locations.

\section{Hyperparameters}
\label{appx: hyperparameters}
For the architecture of AFDT, we follow the default hyperparameters of DT. 
In detail, we use three transformer blocks for most of the environments and one for ball maze environments. 
Each block has one attention head. The embedding dimension is set to 128. 
Dropout rate is set to 0.1. 
We train AFDT for 50000 gradient steps and selected the best checkpoint from 3000 steps, 5000 steps, 10000 steps, 15000 steps, 30000 steps and 50000 steps.
The implementation of AFDT is based on the repository `minimal decision transformer'\cite{minimal_decision_transformer}.
For the architecture of Guided SAC,
the environment Q function, the guided Q function, and the policy net are all three-layer MLPs with ReLU activation function and 256 hidden dimensions.
The implementation of Guided SAC is based on the repository `Stable Baseline 3'\cite{stable-baselines3}.

\section{Action-Free Trajectory Transformer}
\label{appx:aftt}
Our implementation of AFTT is based on the repository `faster-trajectory-transformer' \cite{faster_trajectory_transformer}, which has an improved inference speed compared to the original implementation.
We follow all the default hyperparameters and model architectures in the repository, but remove the action-related component in TT to build AFTT.
We use uniform strategy to discretize the state space.


\end{document}